\documentclass[letterpaper]{article} 
\usepackage[preprint]{aaai2027}  
\usepackage[hyphens]{url}  
\usepackage{graphicx} 
\usepackage{natbib}  
\usepackage{caption} 
\usepackage{algorithm}
\usepackage{algorithmic}

\usepackage{amsmath}
\usepackage{amssymb}
\usepackage{mathtools}
\usepackage{amsthm}
\usepackage{multirow}
\usepackage{mathrsfs}
\usepackage{newfloat}
\usepackage{listings}
\DeclareCaptionStyle{ruled}{labelfont=normalfont,labelsep=colon,strut=off} 
\floatstyle{ruled}
\newfloat{listing}{tb}{lst}{}
\floatname{listing}{Listing}

\usepackage{booktabs}

\definecolor{darkblue}{RGB}{0,0,120}
\definecolor{darkgreen}{RGB}{0,100,0}

\title{Fast Preemptive Robustification: \\High-Frequency Response Anti-Aligns Shared Vulnerability}
\author{
Jiaming Liang, Chi-Man Pun\corresponding
}
\affiliations{
University of Macau, Macau, China\\
chinaliangjm@gmail.com, cmpun@um.edu.mo

}

\begin{document}

\maketitle

\begin{abstract}
Adversarial attacks can readily compromise deep neural networks (DNNs). In particular, transferable attacks (TAs) exploit the shared vulnerabilities among DNNs, enabling perturbations crafted on surrogates to transfer to unseen models. Training-time and post-attack defenses have been extensively studied for combating TAs. Orthogonal to these approaches, preemptive robustification (PR) has emerged as a pre-attack defense that enhances the robustness of benign samples by superimposing protective variations before attacks. Despite its promise, PR remains underexplored and faces several important limitations. First, dependence on well-trained surrogate classifiers limits applicability, as surrogates are task-specific and may even be unavailable in some practical settings. Second, the required iterative optimization or dedicated PR generator training incurs substantial costs. Third, the generated variations are opaque to humans. To address these, we seek an efficient PR that is surrogate-free, optimization-free, training-free, and human-interpretable. Intriguingly, we discover a numerical correlation between the shared vulnerabilities of DNNs and Laplacian responses, with their cosine similarity being significantly negative. This indicates that negated high-frequency response constitutes an important component of shared vulnerabilities. Consequently, strengthening Laplacian responses counteracts this component, improving resistance to TAs. Building upon this insight, we propose \textbf{F}ast \textbf{P}reemptive \textbf{R}obustification (\textbf{FPR}), which performs Laplacian sharpening via a single channel-wise convolution with a $3\times3$ kernel. FPR is simple yet effective, as demonstrated by extensive experiments. Specifically, FPR reduces the attack success rate (ASR) of untargeted TAs by $12.7\%$ and that of targeted TAs from $10.7\%$ to $4.1\%$. Code will be released publicly.

\end{abstract}


\section{Introduction}
\begin{quote}
    \textit{Poisons and medicine are oftentimes the same substance given with different intents.} -Peter Mere Latham
\end{quote}

\noindent\textbf{Background.} Deep neural networks (DNNs)~\cite{szegedy2015going, he2016deep, vaswani2017attention} have revolutionized numerous applications~\cite{kaltenecker2026deep, li2026composable}. Unfortunately, they remain highly vulnerable to adversarial attacks due to high-dimensional non-robust representations~\cite{szegedy2013intriguing, carlini2017towards}. In particular, transferable attacks (TAs)~\cite{goodfellow2015explaining, liu2017delving, dong2018boosting, xie2019improving} approximate shared vulnerabilities across DNNs, enabling adversarial perturbations crafted on surrogate models to effectively transfer to unseen models, thereby hindering reliable DNN deployment even in black-box settings. Mitigating TAs has therefore become a critical challenge.

\begin{figure}[t]
    \centering
    \includegraphics[width=0.95\linewidth]{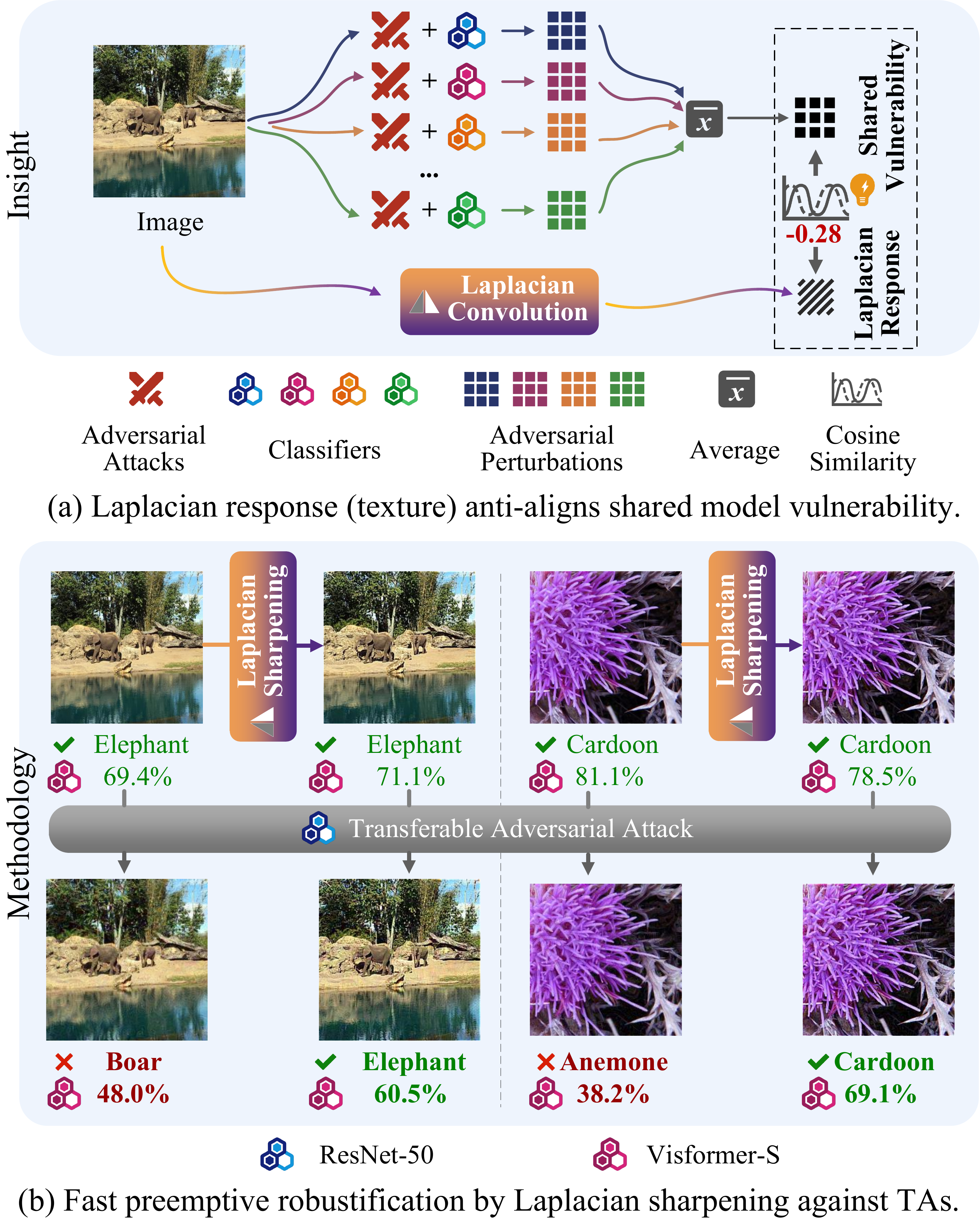}
    \caption{Illustration of the numerical relationship between Laplacian response and shared vulnerability, and the robustness improvement induced by Laplacian sharpening.}
    \label{figure:teaser}
\end{figure}

To address this, training-time and test-time defenses have been extensively studied. Training-time defenses include adversarial training~\cite{sanyal2026antidote}, robust architecture design~\cite{niu2024search}, and robustness-oriented regularization~\cite{chen2026data}. Post-attack defenses mitigate adversarial effects through input transformations~\cite{liang2025comprehensive}, input purification~\cite{jiang2026purmm}, or output correction~\cite{shanmugam2025test}, or identify malevolence via adversarial detection~\cite{fares2026mirrorcheck}. However, both training-time and post-attack defenses typically incur substantial training or inference overhead. More importantly, these approaches require significant modifications to the parameters, architectures, or pipelines of original models, which may introduce undesirable changes and considerably compromise clean accuracy.

Orthogonal to training-time and post-attack defenses, recent preemptive robustification (PR) approaches~\cite{moon2022preemptive, frosio2023best} introduce a pre-attack paradigm that lies between the two. Specifically, PR superimposes small protective variations onto benign samples, making them more resistent to attacks. Instead of imposing robustness requirements on models, PR shifts the burden to samples. This enables defenders to achieve sample-tailored robustness rather than requiring a single model to maintain robustness across diverse inputs, while providing an alternative defense and alleviating training and inference overhead.



\noindent\textbf{Challenges.} However, as an emerging paradigm, PR remains underexplored and faces several challenges. (1) Existing PR methods are tightly coupled with well-trained surrogate classifiers, limiting their applicability in scenarios where such surrogates are unavailable and hindering adaptation to new tasks. (2) Existing methods require iterative optimization or dedicated PR generator training, resulting in substantial computational overhead, particularly for real-time applications. (3) Protective variations generated by existing methods are opaque, raising concerns about their reliability. Therefore, we seek to develop an efficient PR method that is highly effective against TAs, while being surrogate-free, optimization-free, training-free, and human-interpretable.

\noindent\textbf{Insights.} To defend against TAs, an ideal protective variation should neutralize the shared vulnerability among classifiers. Alternatively, a relaxed objective is to identify the dominant components contributing to the shared vulnerability and aggregate their negative counterparts as protective variations.

Prior studies suggest this critical component. (1) Extensive studies~\cite{yin2019fourier, wang2020high, caro2020local, pei2025diffusion} have shown that adversarial perturbations often exhibit dominant energy in the mid-to-high frequency spectrum. (2) Protective variations generated by previous PR methods~\cite{frosio2023best} exhibit texture-like patterns, and the robustified samples resemble texture- and contrast-enhanced images. (3) The attackability measure OTI~\cite{liang2026oti} reveals that images with stronger textures tend to be more resistant to attacks.

Collectively, these findings lead us to a hypothesis: \textcolor{darkblue}{the negated high-frequency response of an image constitutes a critical component of the shared vulnerability on the image across models.} Our experiments (see Section~\ref{section: Laplacian Response}) quantitatively validate this hypothesis: \textcolor{darkblue}{the mean iterative white-box perturbation across multiple classifiers on the same sample exhibits a significantly negative cosine similarity with the sample's Laplacian response, and its magnitude increases with more classifiers and iterations}, as shown in Figure~\ref{figure:teaser}(a).


\noindent\textbf{Methodology.} Therefore, we can achieve PR by enhancing the high-frequency response of images to counteract this critical component of shared vulnerabilities. With an appropriate magnitude, Laplacian sharpening is not only imperceptible to humans but also achieves strong PR performance. Therefore, we adopt Laplacian sharpening to enhance high-frequency responses as a PR strategy (see Section~\ref{section: Fast Preemptive Robustification}). Laplacian sharpening requires only a single channel-wise $3\times3$ convolution, making it highly efficient even on commodity CPUs. Accordingly, we refer to this method as \textbf{F}ast \textbf{P}reemptive \textbf{R}obustification (\textbf{FPR}). 

Extensive experiments (see Section~\ref{section: Experiments and Results}) demonstrate the effectiveness of FPR. FPR has a negligible impact on benign accuracy while substantially improving robustness against TAs. On average, FPR reduces the attack success rate (ASR) of single-surrogate untargeted TAs by $12.7\%$, ensemble-surrogate untargeted ASR by $14.7\%$, and targeted ASR from $10.7\%$ to $4.1\%$. Moreover, FPR is complementary to training-time and test-time defenses. For example, when combined with state-of-the-art adversarial training methods, FPR further reduces the ASR from $12.1\%$ to $10.3\%$.

It is remarkable that such a classical image enhancement technique can deliver substantial benefits in the deep learning era. The main contributions are summarized as follows:
\begin{itemize}
    \item To the best of our knowledge, this is the first work to reveal a quantitative similarity between an image's high-frequency response and shared model vulnerabilities.

    \item Based on this similarity, we develop FPR with Laplacian sharpening. FPR is the first surrogate-free, optimization-free, training-free, and human-interpretable PR method.

    \item Extensive experiments demonstrate that FPR achieves strong robustness against diverse TAs while incurring negligible computational costs, enabling real-time PR.



\end{itemize}

\section{Related Work}

\subsection{Transferable Attacks}
In the real world, attackers typically lack access to the architecture and parameters of the victim model, giving rise to the black-box threat model. Two mainstream paradigms, query-based attacks and transferable attacks, have been developed for black-box attacks. Query-based attacks refine perturbations by exploiting the feedback from the victim model. In contrast, transferable attacks enhance the adversarial transferability so that perturbations crafted on surrogate models remain effective against unseen models. 
Diverse strategies have been proposed to improve transferability, including gradient-based~\cite{gan2025boosting, peng2025boosting}, model-related~\cite{wang2025improving, liu2026harnessing}, transformation-based~\cite{liang2025ic, guo2025boosting}, advanced objective~\cite{zheng2025enhancing, liupixel2feature}, generative~\cite{nakka2025nat, li2025aim}, and ensemble-based~\cite{chen2023rethinking, tang2024ensemble} approaches.

\subsection{Adversarial Defenses}
In response to the growing adversarial threat, various defenses have been developed. Depending on the deployment stage, these defenses can be classified as training-time defenses, post-attack defenses, and recent pre-attack defenses.

\textbf{Training-time defenses} enhance model robustness during the training phase. Adversarial training~\cite{zhang2025adversarial, sanyal2026antidote} does so by incorporating adversarial examples into the training set. Robust architecture design~\cite{peng2023robust, niu2024search} leverages structural designs to mitigate adversarial effects. Robustness-oriented regularization~\cite{zuhlke2025adversarial, chen2026data} leverages tailored regularization terms.

\textbf{Post-attack defenses} alleviate adversarial threats by mitigating or detecting adversarial effects after attacks. Input purification defenses~\cite{yang2024adversarial, lei2025instant} adjust adversarial examples toward the benign data distribution via generative methods such as GANs and diffusion models. Input transformation defenses~\cite{cohen2024simple, liang2025comprehensive} transform inputs to disrupt the structure of adversarial perturbations. Adversarial detection~\cite{liu2024lightweight, kong2025data} prevents malicious access by identifying adversarial effects in the input. Output correction defenses~\cite{shanmugam2025test} aggregate multiple prediction copies to produce more reliable outputs.

\textbf{Pre-attack defenses.} Beyond these two stages, there exists another stage after model deployment but before inference, which is responsible for sample storage and transmission. We term defenses operating at this stage as pre-attack defenses. Unlike training-time and post-attack defenses, which primarily improve robustness from a model-centric perspective, pre-attack defenses shift the focus toward enhancing the intrinsic robustness of the samples themselves. 

Specifically, preemptive robustification superimposes protective variations onto samples, making them more resistant to adversarial attacks. Unfortunately, existing PR methods remain scarce. Bi-Level~\cite{moon2022preemptive} preemptively robustifies images by solving a bi-level optimization problem to find nearby robust points. $A^{5}$~\cite{frosio2023best} leverages automatic perturbation analysis to craft certified defensive perturbations against bounded attacks. Despite these advances, as discussed above, their limitations hinder practical deployment. Therefore, this work aims to develop an efficient PR method to overcome their limitations.

\subsection{Image Sharpening in Deep Learning Security}
Traditionally, image sharpening has been widely studied as a means of enhancing perceptual quality. Such approaches range from classical spatial-domain and frequency-domain filtering~\cite{gonzalez2009digital} to adaptive filtering~\cite{schavemaker2000image, zhang2008adaptive} methods, and more recently, DNN-based sharpening techniques~\cite{xu2025laboring}. In recent years, image sharpening has found renewed applications in the deep learning era. It has been explored as an augmentation strategy to improve model generalization and robustness~\cite{he2023shift, he2023edge, heinert2024reducing}, and as a post-attack defense to mitigate adversarial effects~\cite{mustafa2019image, he2023edge}. These studies suggest that image sharpening, beyond its traditional role in perceptual enhancement, can also contribute to deep learning security. In this work, we revisit image sharpening from the pre-attack defense perspective and uncover its potential for PR.


\section{Theory and Methodology}

\subsection{Preliminaries}
\label{section: Preliminaries}
\noindent\textbf{Transferable Attacks.} Given a benign sample set $\mathcal{X}=\{\boldsymbol{x}_{i}\}_{i=1}^{N}$, a TA $A\in\mathcal{A}$ crafts highly transferable perturbations on a surrogate $f_{\mathrm{s}}$, yielding the adversarial example set:
\begin{equation}
\begin{aligned}
\mathcal{X}_{\mathrm{adv}}=\large\{\boldsymbol{x}+\boldsymbol{\delta}_{\mathrm{adv}}\mid \boldsymbol{\delta}_{\mathrm{adv}}=A\large(\boldsymbol{x};f_{\mathrm{s}}\large),\boldsymbol{x}\in \mathcal{X}\large\},
\end{aligned}
\end{equation}
such that unseen classifiers $\mathcal{F}=\{f_{j}\}_{j=1}^{N}$ can be severely misled, and the attack success rate (ASR) are calculated by
\begin{equation}
\begin{aligned}
\mathrm{ASR}\large(\mathcal{X},A,f_{\mathrm{s}},\mathcal{F}\large)=\mathbb{E}_{\boldsymbol{x}, f}\mathbf{1}\large\{f\large(\boldsymbol{x}+\boldsymbol{\delta}_{\mathrm{adv}}\large)\neq  y\large\},
\end{aligned}
\end{equation}
where $y$ is the true label of corresponding benign sample $\boldsymbol{x}$ and $\mathbf{1}\{\cdot\}$ is the indicator function. Here, the adversarial perturbations are constrained by $\ell_p$-norm attack budget $\epsilon_{\mathrm{adv}}$:
\begin{equation}
\begin{aligned}
\|\boldsymbol{\delta}_{\mathrm{adv}}\|_p\leq \epsilon_{\mathrm{adv}}.
\end{aligned}
\end{equation}
Following previous work, we primarily consider $p=\infty$.

\noindent\textbf{Preemptive Robustification.} A PR method $R\in\mathcal{R}$ aims to generate a protective variation $\boldsymbol{\delta}_{\mathrm{r}}$ for each benign sample $\boldsymbol{x}$, yielding a preemptively robustified sample set:
\begin{equation}
\begin{aligned}
\mathcal{X}_{\mathrm{r}}=\large\{\boldsymbol{x}+\boldsymbol{\delta}_{\mathrm{r}}\mid \boldsymbol{\delta}_{\mathrm{r}}=R\large(\boldsymbol{x}\large),\boldsymbol{x}\in \mathcal{X}\large\},
\end{aligned}
\end{equation}
that is more resistant to TAs. The gains are measured by:
\begin{equation}
\begin{aligned}
\Delta=\mathrm{ASR}\large(\mathcal{X},A,f_{\mathrm{s}},\mathcal{F}\large)-\mathrm{ASR}\large(\mathcal{X}_{\mathrm{r}},A,f_{\mathrm{s}},\mathcal{F}\large).
\end{aligned}
\end{equation}

\subsection{Laplacian Response and Shared Vulnerability}
\label{section: Laplacian Response}
We begin with an idealized setting. Suppose the defender has full knowledge of the classifier set $\mathcal{F}$ to be attacked. For a benign sample $\boldsymbol{x}$, the ideal protective variation $\boldsymbol{\delta}_{\mathrm{r}}^{*}$ should correspond to the opposite of the expected vulnerability of classifiers in $\mathcal{F}$ with respect to $\boldsymbol{x}$. We characterize the vulnerability of a classifier on $\boldsymbol{x}$ by its white-box perturbation. Accordingly, the ideal $\boldsymbol{\delta}_{\mathrm{r}}^{*}$ can be formulated as:
\begin{equation}
\begin{aligned}
\boldsymbol{\delta}_{\mathrm{r}}^{*}=-\mathbb{E}_{f\in\mathcal{F}}\large[A\large(\boldsymbol{x},f\large)\large].
\label{equation:ideal_protective_variation}
\end{aligned}
\end{equation}
However, in practice, the defender typically lacks prior knowledge of the classifiers that will process the stored or transmitted sample $\boldsymbol{x}$, as well as the potential downstream tasks involved. Moreover, as discussed above, PR methods that rely on surrogate models and optimization or training procedures suffer from several limitations. Therefore, Equation~\ref{equation:ideal_protective_variation} is clearly difficult to achieve in real world.

\begin{figure*}[t]
    \centering
    \includegraphics[width=0.99\linewidth]{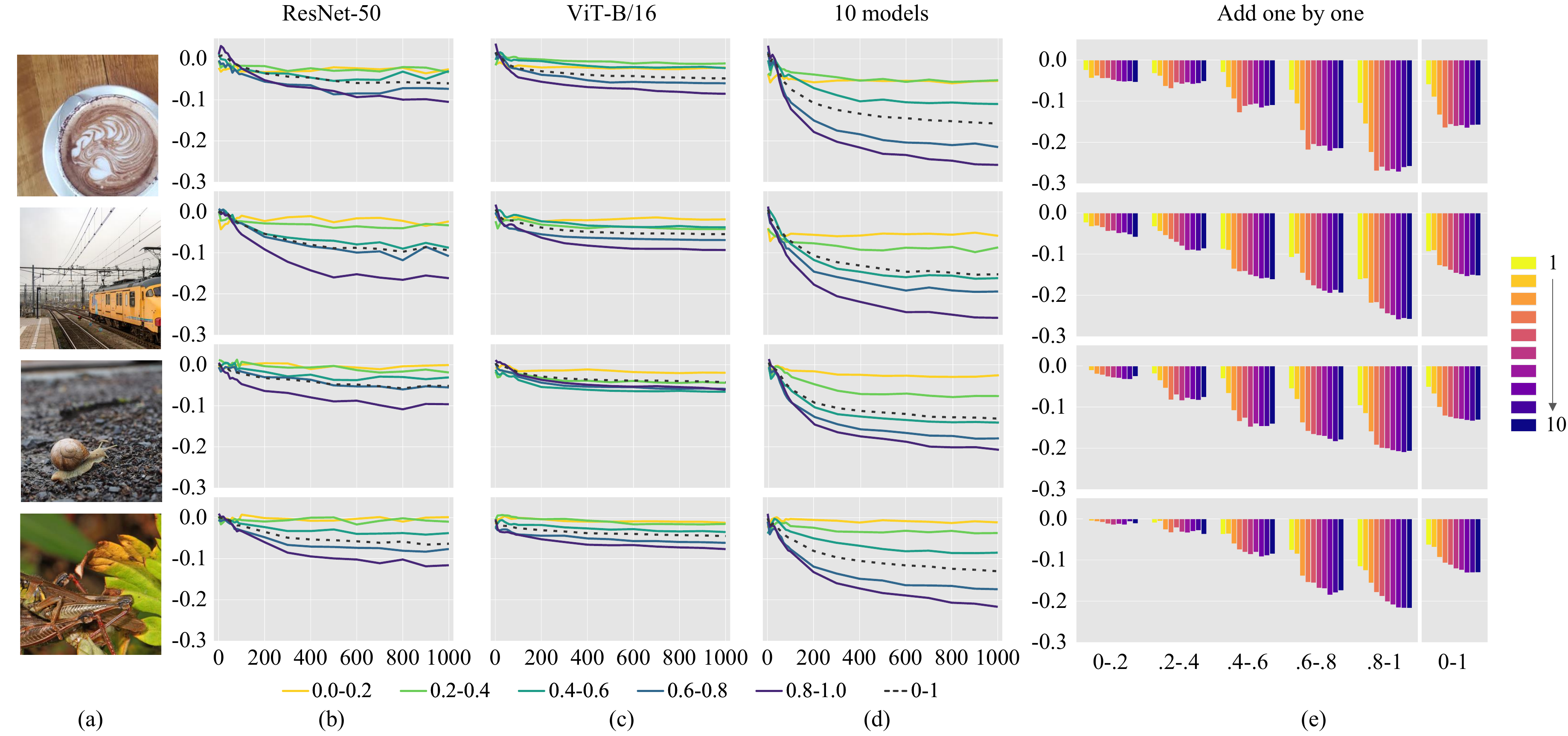}
    \caption{Cosine similarities between image Laplacian responses and their corresponding adversarial perturbations (generated by MI-FGSM). (a) Original images. (b) Perturbations generated on ResNet-50. X-axis denotes the attack iteration and y-axis denotes the cosine similarity. Different colored curves represent the cosine similarities within different texture intensity intervals. For example, $0.6-0.8$ indicates the cosine similarity calculated from texture elements ranked within the $60\%-80\%$ intensity range after ascending sorting. (c) Perturbations generated on ViT-B/16. (d) Perturbations generated from 10 different surrogate models are averaged before computing cosine similarities. (e) The x-axis denotes texture intensity intervals, and bars with different colors indicate the number of surrogate models. The iteration number is $1000$.}
    \label{fig:cossim}
\end{figure*}

Although the ideal protective variation $\boldsymbol{\delta}_{\mathrm{r}}^{*}$ is inaccessible in deployment, we can instead analyze its critical components. By taking the negation of such components as the approximation $\hat{\boldsymbol{\delta}}_{\mathrm{r}}$, we can partially compensate for the shared vulnerability and thereby improve resistance against TAs. Therefore, we shift our focus to analyzing the components underlying shared vulnerability.


We seek potential clues from existing studies. First, extensive research~\cite{yin2019fourier, wang2020high, caro2020local, pei2025diffusion} has shown that the energy of adversarial perturbations is concentrated in the mid-to-high frequency regions. Second, robustified samples generated by previous PR methods~\cite{moon2022preemptive, frosio2023best} consistently exhibit stronger textures and enhanced contrast compared with the original images. Furthermore, recent image attackability measure OTI~\cite{liang2026oti} has revealed that samples with stronger textures tend to be more resistant to adversarial attacks. Collectively, these findings from spectral analysis, empirical visualization of PR methods, and sample-centric robustness analysis all point to a close relationship between image high-frequency responses and adversarial perturbations. This naturally raises the question of whether high-frequency responses are closely associated with shared vulnerability, or even constitute the critical component of the shared vulnerability.

\begin{figure}[t]
    \centering
    \includegraphics[width=0.99\linewidth]{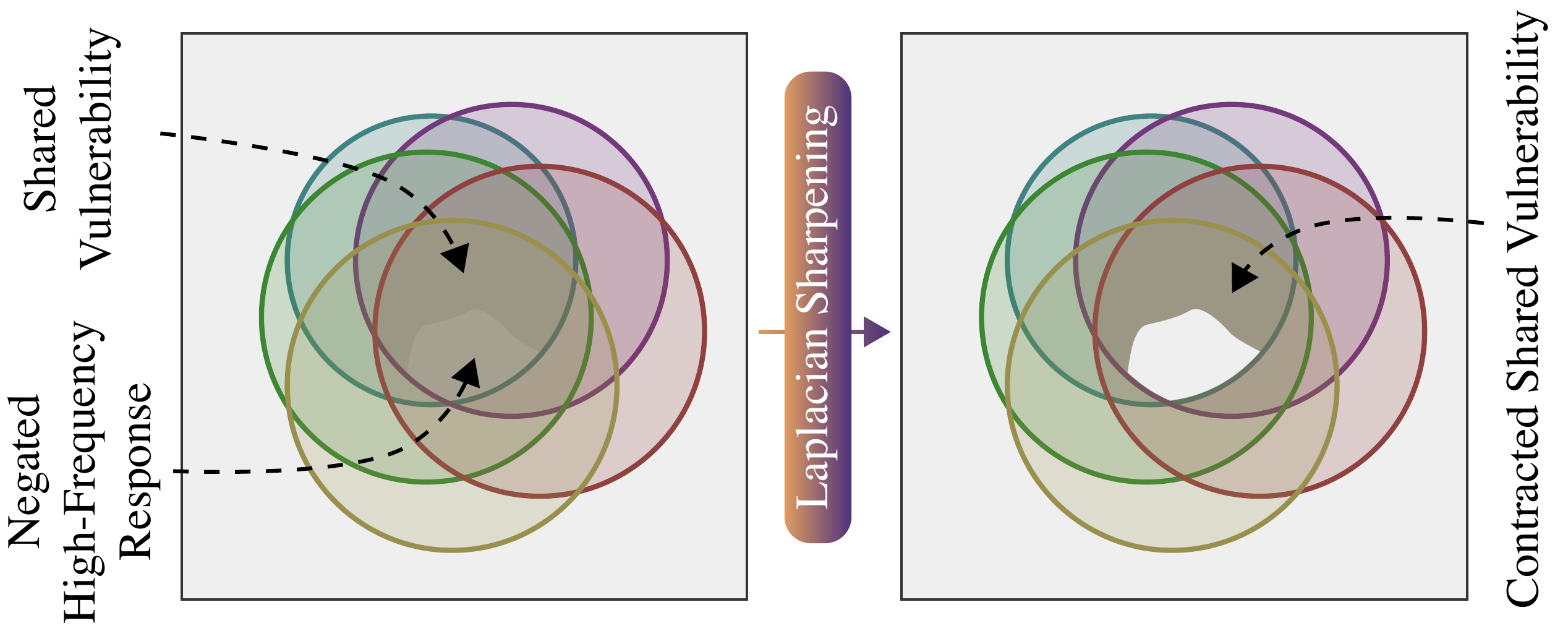}
    \caption{Illustration of Laplacian sharpening in contracting shared vulnerability by enhancing high-frequency response. Different circles represent the vulnerabilities of different models to the same image.}
    \label{fig:explanation}
\end{figure}

To this end, we design a dedicated experiment to investigate the quantitative relationship between high-frequency responses and shared vulnerability. Specifically, we convolve the image with the 8-neighborhood Laplacian kernel $K_{\mathscr{L}}$ to extract the high-frequency response: 
\begin{equation}
\begin{aligned}
    \boldsymbol{x}_{\mathscr{L}} = K_{\mathscr{L}} * \boldsymbol{x}.
\end{aligned}
\end{equation}
We generate $\ell_{\infty}$-bounded white-box adversarial perturbations (based on MI-FGSM) across ten classifiers, including R50, IncV3, D121, EB2, ConvN-B, ViT-B/16, BeiT-B/16, Swin-B, CaF, and PoolF. We then average these perturbations to obtain $\bar{\boldsymbol{\delta}}_{\mathrm{adv}}$ (without averaging when only one classifier is used), which serves as an empirical approximation of the shared vulnerability. Finally, the cosine similarity $\eta$ is:
\begin{equation}
\begin{aligned}
    \eta = \frac{\boldsymbol{x}_{\mathscr{L}}\cdot\bar{\boldsymbol{\delta}}_{\mathrm{adv}}}{\|\boldsymbol{x}_{\mathscr{L}}\|\cdot\|\bar{\boldsymbol{\delta}}_{\mathrm{adv}}\|}.
\end{aligned}
\end{equation}
The illustrated numerical results are presented in Figure~\ref{fig:cossim}.

\begin{figure*}[t]
    \centering
    \includegraphics[width=0.92\linewidth]{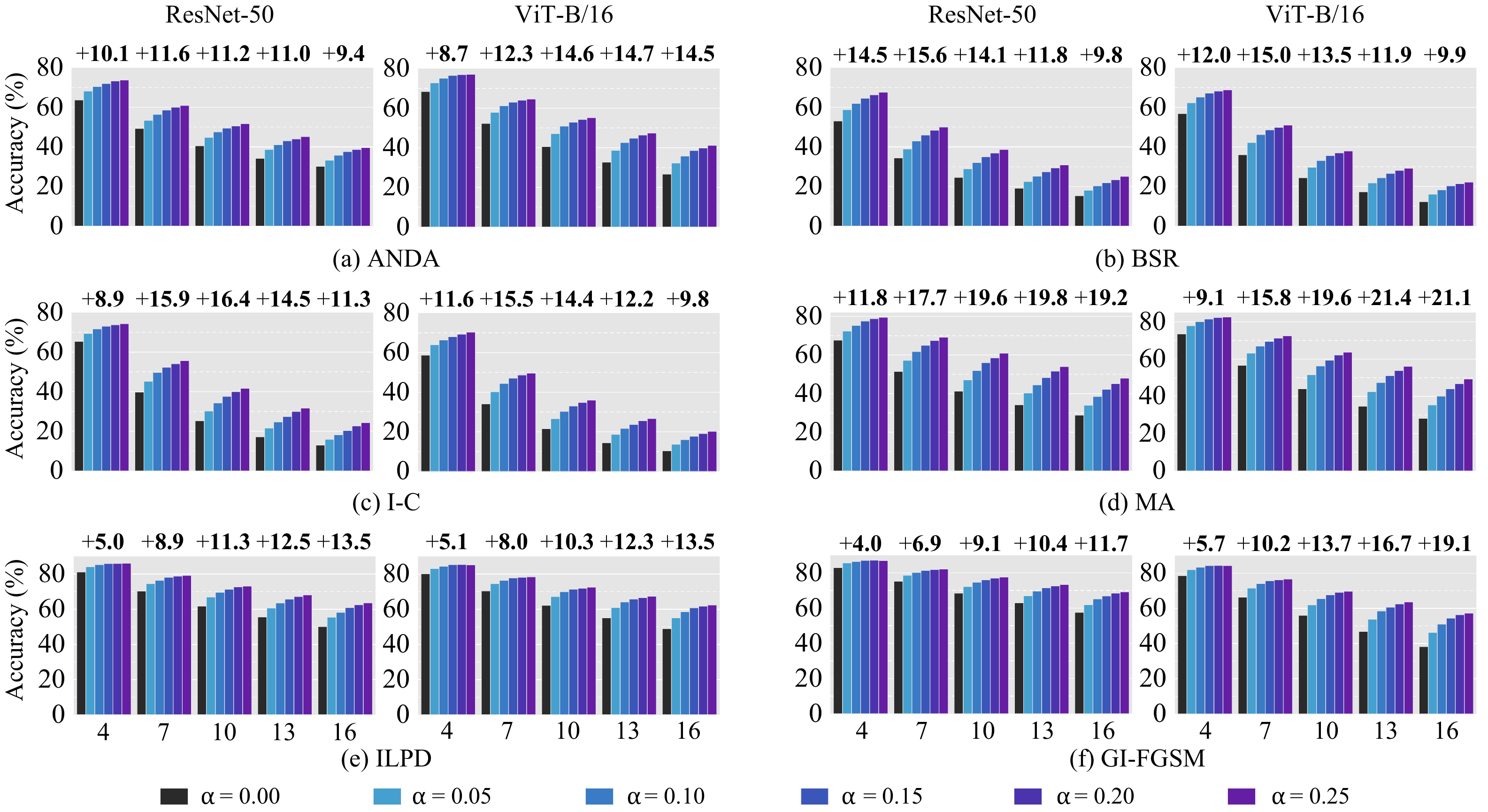}
    \caption{Accuracy (\%) under various non-targeted transferable attacks and $\epsilon_{\mathrm{adv}}$. Gains from $\alpha=0.00$ (w/o FPR) to $\alpha=0.25$ are annotated at the top. Similar annotations apply hereafter.}
    \label{fig:nontargeted_blackbox_NIPS17}
\end{figure*}

\begin{figure*}[t]
    \centering
    \includegraphics[width=0.90\linewidth]{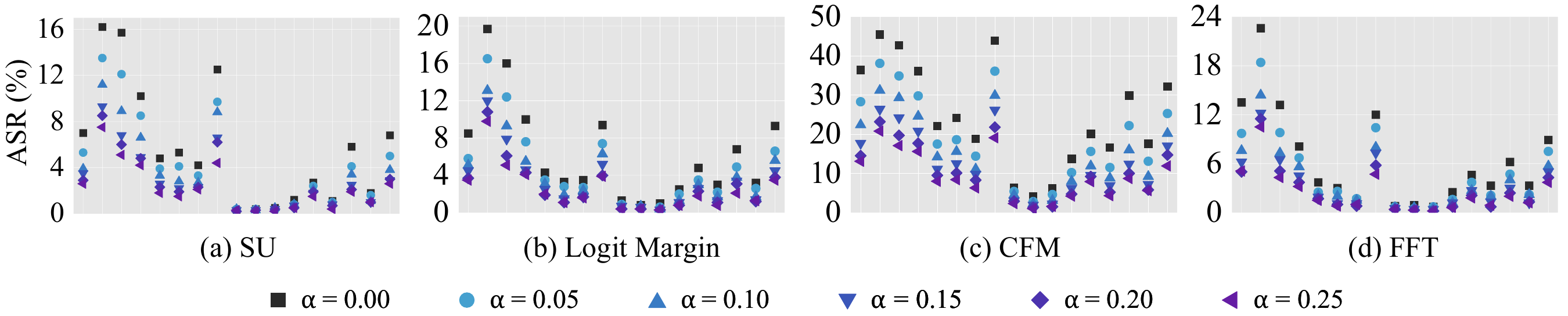}
    \caption{ASRs (\%) of targeted transferable attacks on robustified samples by FPR with varying $\alpha$.}
    \label{fig:targeted_NIPS17}
\end{figure*}

\begin{figure}[t]
    \centering
    \includegraphics[width=0.90\linewidth]{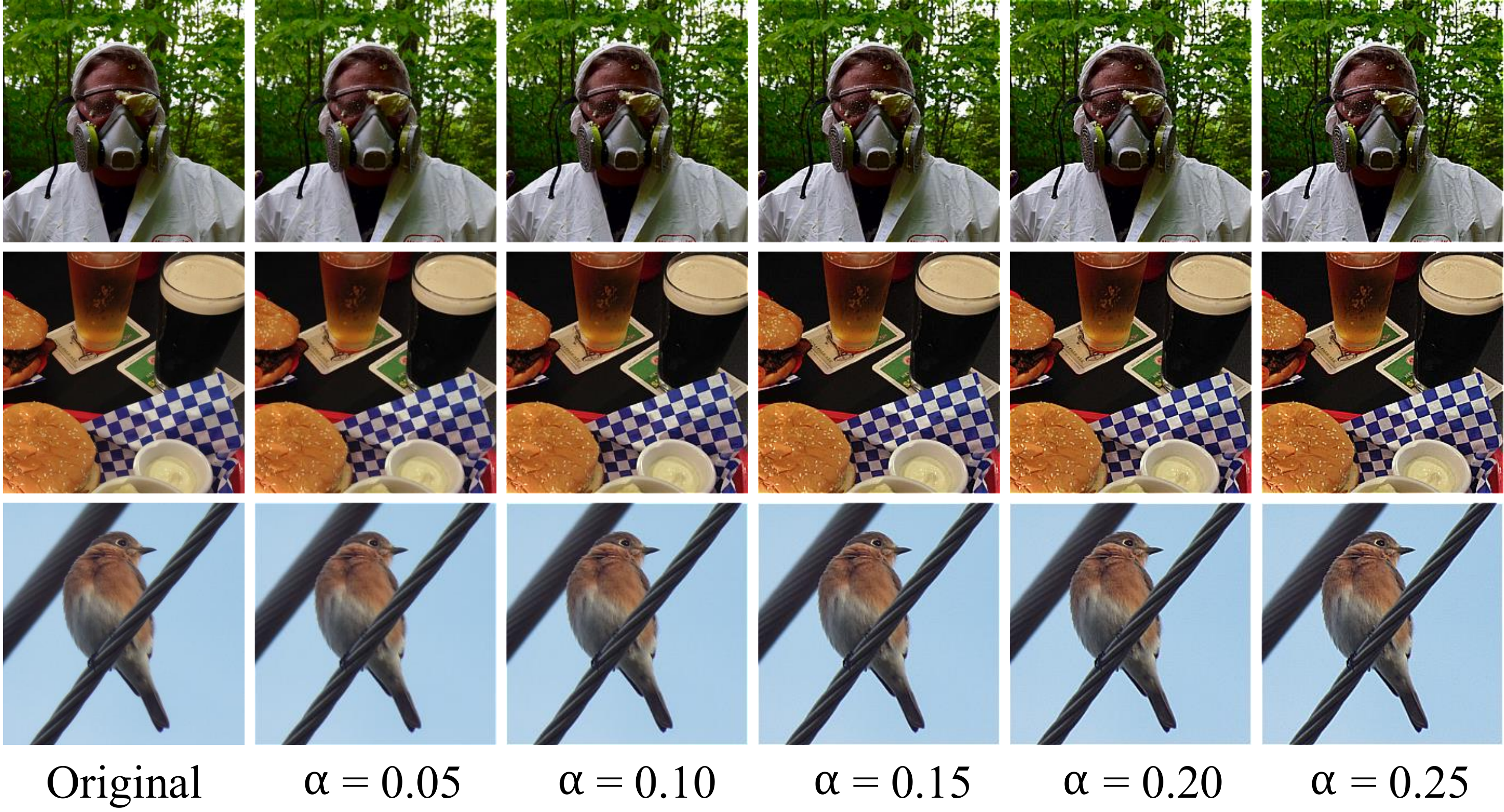}
    \caption{Laplacian sharpened images with varying $\alpha$.}
    \label{figure:examples_of_laplacian_sharpening}
\end{figure}

Since an image $\boldsymbol{x}$ is a high-dimensional tensor (of size $3\times224\times224$ in our experiments), if high-frequency responses were unrelated to shared vulnerability, the expected $\eta$ would be close to zero due to the concentration effect in high-dimensional spaces. As shown in Figure~\ref{fig:cossim}, this is indeed the case when adversarial perturbations are insufficiently optimized (the low-iteration regime in Figure~\ref{fig:cossim}(b–d)) or generated from only a single surrogate model (Figure~\ref{fig:cossim}(e)), where $\eta$ remains close to zero. These results indicate that perturbations obtained before convergence or from a few surrogates exhibit little correlation with the Laplacian response. Although it has long been hypothesized that model vulnerability is related to high-frequency image components, previous studies have struggled to establish a statistically significant quantitative similarity, consistent with these observations.

However, as the optimization proceeds and the number of surrogate models increases, cosine similarity $\eta$ between the mean of converged adversarial perturbations and the Laplacian response steadily rises to approximately $0.2$. This finding is encouraging, as it provides empirical evidence that high-frequency responses constitute a critical component of the shared vulnerability, while also revealing a concrete quantitative similarity between them. Furthermore, we compute the cosine similarity over regions with different Laplacian-response magnitudes and observe that regions with stronger responses exhibit higher similarity. Specifically, the cosine similarity approaches $0.3$ when evaluated on the top $20\%$ elements with the strongest Laplacian response.

\subsection{Fast Preemptive Robustification}
\label{section: Fast Preemptive Robustification}
Therefore, we demonstrate that high-frequency responses constitute an important component of shared vulnerability. By enhancing the high-frequency responses of images, this vulnerable component can be effectively compensated, thereby shrinking the shared vulnerability exploited by TAs and rendering them ineffective, as illustrated in Figure~\ref{fig:explanation}.

\begin{figure*}[t]
    \centering
    \includegraphics[width=0.92\linewidth]{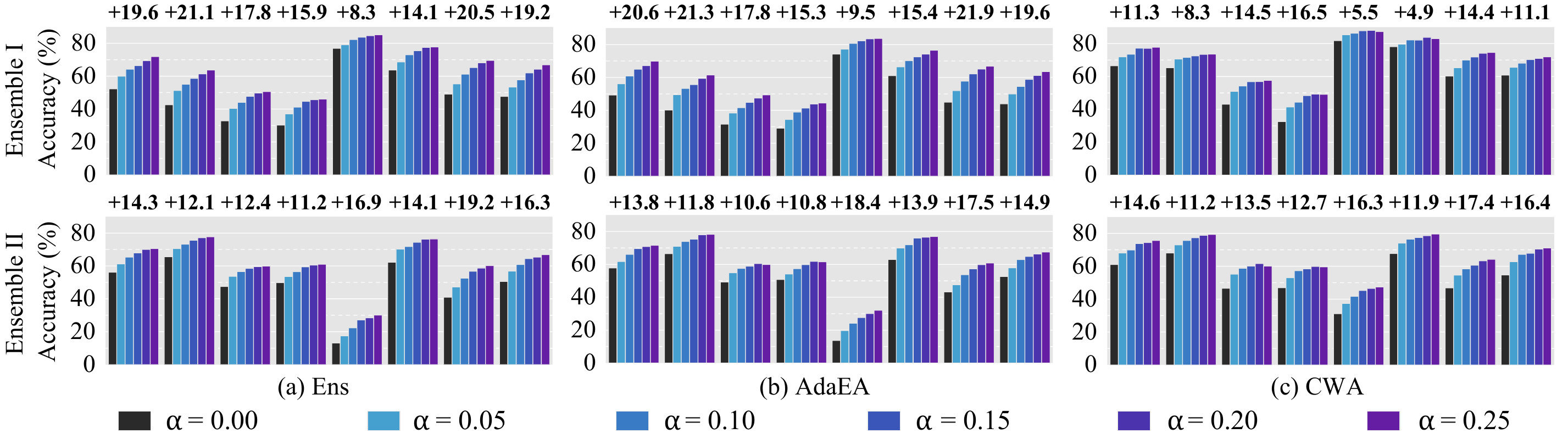}
    \caption{Accuracy (\%) under various ensemble-based attacks, evaluated on robustified images by FPR with different $\alpha$.}
    \label{fig:ensemble_based_attacks_NIPS17}
\end{figure*}

\begin{figure*}[t]
    \centering
    \includegraphics[width=0.80\linewidth]{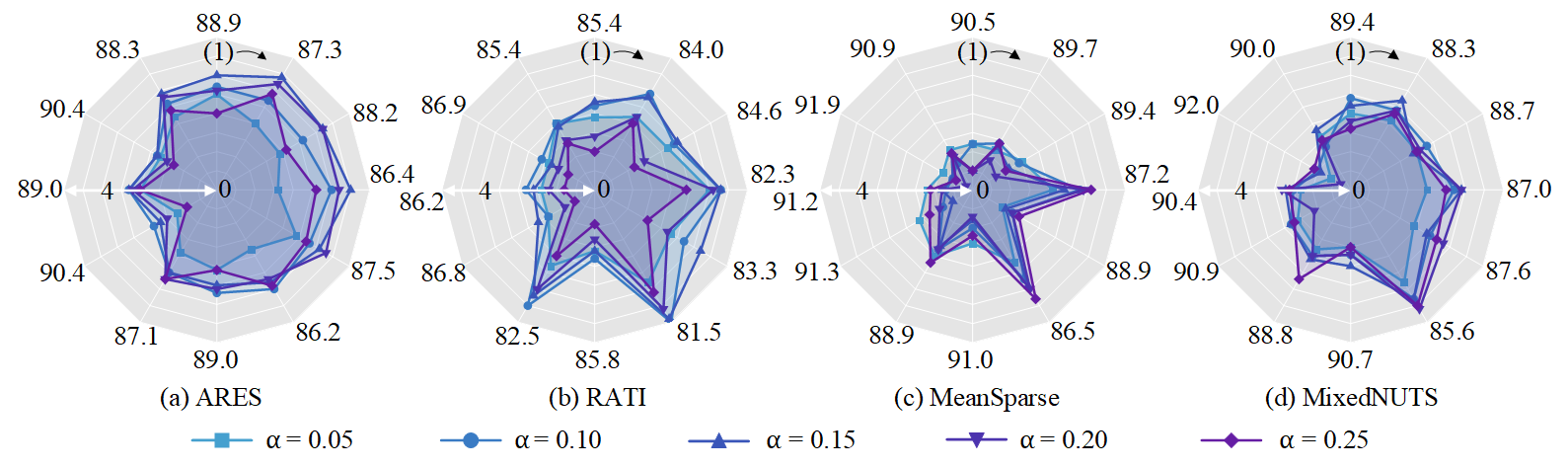}
    \caption{Gains (\%) of adversarially trained models integrated with our FPR. The axes represent: ANDA (1–2), BSR (3–4), I-C (5–6), MA (7–8), ILPD (9–10), and GI-FGSM (11–12). Odd indices use R50 as the surrogate, while even indices use VB16.}
    \label{fig:nontargeted_adversarially_trained_NIPS17}
\end{figure*}

To this end, we introduce Laplacian sharpening as a novel PR method to defend against TAs, where the strength is controlled by the sharpening coefficient $\alpha (\alpha>0)$. This method can be formulated as follows:
\begin{equation}
\begin{aligned}
\boldsymbol{x}_{r}=\mathrm{Clip}_{[0,1]}\big(\boldsymbol{x}+\alpha\cdot\large(K_{\mathscr{L}}\ast \boldsymbol{x}\large)\big).
\end{aligned}
\label{equation:definition_Laplacian_sharpening}
\end{equation}
Although the proposed method is extremely simple, requiring only a single channel-wise convolution with a $3\times3$ kernel, the experiments in Section~\ref{section: Experiments and Results} demonstrate its effectiveness. In addition, this method inherits the favorable properties of Laplacian sharpening. With a small $\alpha$, the robustified images are imperceptible to humans, as shown in Figure~\ref{figure:examples_of_laplacian_sharpening}. Even when the enhancement is slightly noticeable, Laplacian sharpening itself is a technique designed to improve perceptual quality. Unlike optimization-based or generator-based approaches, it does not compromise the visual quality. 

We name this method \textbf{F}ast \textbf{P}reemptive \textbf{R}obustification (\textbf{FPR}), which is the first surrogate-free, optimization-free, training-free, and human-interpretable PR method.

\section{Experiments and Results}
\label{section: Experiments and Results}
\subsection{Setup}
\label{section: Setup}
\textbf{Benchmarks.} We conduct a comprehensive evaluation across a broad suite of attack benchmarks: ANDA~\cite{fang2024strong} and GI-FGSM~\cite{wang2024boostingo} for gradient-based attacks, BSR~\cite{wang2024boosting} and I-C~\cite{liang2025ic} for transformation-based attacks, ILPD~\cite{li2023improving} for advanced objective attack, MA~\cite{ma2024improving} for model-related attacks. The targeted attacks include SU~\cite{wei2023enhancing}, Logit Margin~\cite{weng2023logit}, CFM~\cite{byun2023introducing}, and FFT~\cite{zeng2024enhancing}. The ensemble-based attacks include Ens~\cite{liu2017delving}, AdaEA~\cite{chen2023adaptive}, and CWA~\cite{chen2024rethinking}. 

\noindent\textbf{Datasets.} Following previous work, our experiments are conducted on the NIPS 2017 Adversarial Competition dataset~\footnote{\url{https://github.com/anlthms/nips-2017.git}}.

\noindent\textbf{Models.} The models are listed as follows: (1) ResNet-50 (R50), (2) ConvNeXt-B (ConvN), (3) WideResNet-50 (WR50), (4) DenseNet-161 (D161), (5) RegNet-X (RegN), (6) EfficientNet-B2, (7) GoogLeNet (GgN), (8) Xception-71 (Xcept), (9) Inception-ResNet-V2 (IR2), (10) ViT-B/16 (VB16), (11) Convformer (ConvF), (12) Visformer (VisF), (13) Poolformer (PoolF), (14) ViT-B/8 (VB8), (15) Swin Transformer-B (Swin), (16) PiT-B (PiT), (17) XCiT-S (XCiT), and (18) Caformer-M36 (CaF). 


\noindent\textbf{Environment.} All experiments are conducted on a single NVIDIA A100 GPU, with the random seed set to 0.

\begin{figure*}[t]
    \centering
    \includegraphics[width=0.92\linewidth]{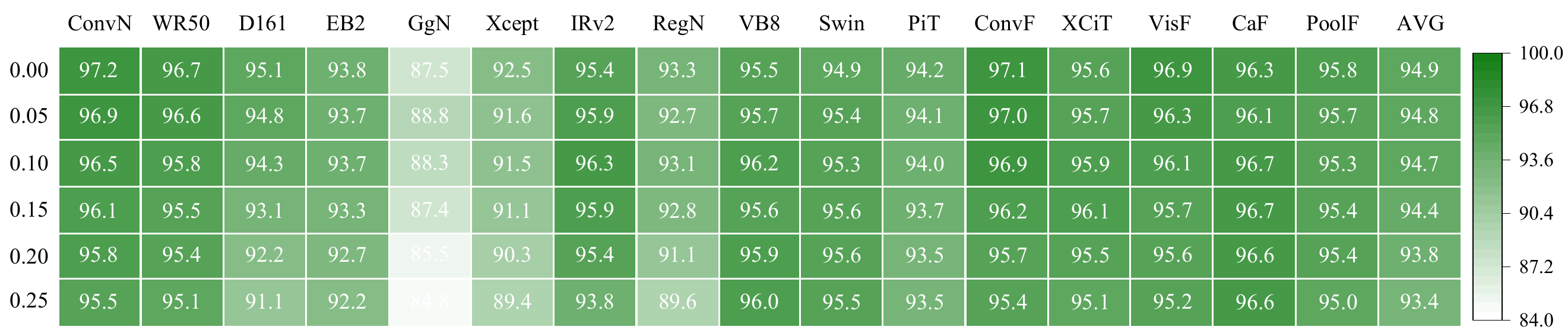}
    \caption{Impact of FPR with varying $\alpha$ on standard accuracy (\%).}
    \label{fig:impact_on_standard_classification_accuracy}
\end{figure*}

\begin{figure*}[t]
    \centering
    \includegraphics[width=0.92\linewidth]{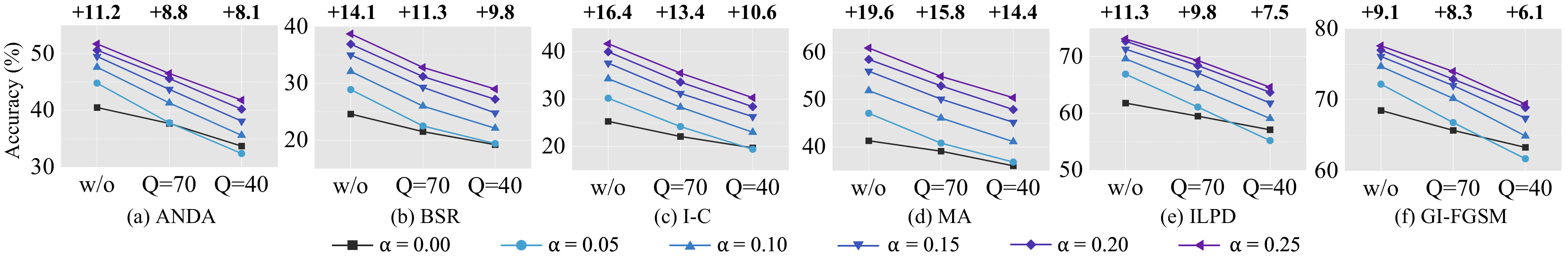}
    \caption{Impact of JPEG compression on our FPR.}
    \label{fig:experiments_JPEG}
\end{figure*}

\subsection{To Non-Targeted Transferable Attacks}
\label{section: Non-Targeted Attacks on Classification}
First, we evaluate FPR with different coefficients $\alpha$ in defending against diverse advanced non-targeted attacks with varying perturbation budgets $\epsilon_{\mathrm{adv}}$. The perturbation budget $\epsilon_{\mathrm{adv}}$ spans $\{4, 7, 10, 13, 16\}/255$ of $\ell_{\infty}$-norm, with the iterations $T=10$ and the step size $\epsilon_{\mathrm{adv}}/T$. The surrogates include R50 and VB16. The target models consist of the remaining 17 models listed in Section~\ref{section: Setup}. Figure~\ref{fig:nontargeted_blackbox_NIPS17} demonstrates that FPR is highly effective against non-targeted transferable attacks. On average, FPR reduces ASR by $12.7\%$.

    
    

\subsection{To Targeted Transferable Attacks}
\label{section: Targeted Attacks on Classification}
In this experiment, we evaluate our FPR against targeted transferable attacks. The target labels are \textit{1} and \textit{500}. The surrogate model is R50. The target models include (2)-(5) and (10)-(13) to which targeted transferable attacks are effective. The budget $\epsilon_{\mathrm{adv}}$ is $16/255$ with the iterations $T=200$ and the step size $\epsilon_{\mathrm{adv}}/T$. Results in Figure~\ref{fig:targeted_NIPS17} indicate that our FPR provides strong robustification against transferable targeted attacks. On average, FPR reduces ASR from $10.7\%$ to $4.1\%$.

\subsection{To Ensemble-based Transferable Attacks}
Beyond evaluating FPR against single-surrogate attacks, we are also interested in the effectiveness against ensemble-based transferable attacks. In this experiment, ensemble I = \{R50, EB0, IncV3\} and ensemble II = \{VS16, VS32, BeiTB16\} are used as surrogates to evaluate across various ensemble-based attacks. The target models are the same as those in Section~\ref{section: Targeted Attacks on Classification}. The budget $\epsilon_{\mathrm{adv}}=10/255$ with $T=10$. Results in Figure~\ref{fig:ensemble_based_attacks_NIPS17} demonstrate that FPR provides a strong defense against ensemble-based attacks. On average, FPR reduces ASR by $14.7\%$.

\subsection{Integrating with Adversarial Training}
Above experiments evaluate FPR in isolation, without combining it with other defenses. Here, we investigate the effect of jointly applying FPR and adversarial training. We test four advanced methods: ARES (Swin-B)~\cite{liu2025comprehensive}, RATI~\cite{singh2023revisiting} (ViT-S + ConvStem), MeanSparse~\cite{amini2024meansparse} (ConvNeXt-L), and MixedNUTS~\cite{bai2024mixednuts} (ConvNeXtV2-L + Swin-L). Adversarial examples are generated as in Section~\ref{section: Non-Targeted Attacks on Classification}. Results in Figure~\ref{fig:nontargeted_adversarially_trained_NIPS17} indicate that FPR can further enhance robustness.

\subsection{Impact on Standard Accuracy}
This experiment evaluates the impact of FPR on benign samples. The results in Figure~\ref{fig:impact_on_standard_classification_accuracy} demonstrate that FPR achieves a favorable trade-off between adversarial robustness and standard accuracy. On average, when $\alpha=0.10$, FPR incurs only a $0.2\%$ accuracy fluctuation. Even with a $\alpha=0.25$, the accuracy fluctuation remains merely $1.5\%$. Compared with other advanced defenses~\cite{wang2026robust, mirsadeghi2026active}, this negligible impact highlights the advantage of FPR.

\subsection{Robustness against JPEG Compression}
Robustified samples may undergo post-processing operations such as compression during storage and transmission. Here, we further investigate whether FPR-robustified samples can maintain robustness after such operations. This experiment considers practical JPEG compression as an example. Specifically, we first apply JPEG compression to the FPR-robustified samples, then generate adversarial examples from the compressed samples and evaluate their robustness. Compression quality factors $Q$ are set to $40$ and $70$. Results presented in Figure~\ref{fig:experiments_JPEG} demonstrate that FPR provides stable robustness gains under such aggressive post-processing.

\section{Conclusions and Limitations}
This work first reveals a strong cosine similarity between image high-frequency responses and the shared model vulnerability. This finding further sheds light on the intrinsic relationship between image itself and perturbations. Building upon this discovery, we propose FPR, the first surrogate-free, optimization-free, training-free, and human-interpretable PR method, which effectively mitigates transferable attacks through conventional Laplacian sharpening. Extensive experiments demonstrate that, despite requiring only a single channel-wise $3\times3$ convolution, FPR achieves favorable robustness-accuracy trade-off. Moreover, FPR exhibits strong resilience against post-processing transformations, highlighting its practical applicability. Nevertheless, this work introduces a novel perspective on PR. We hope it will inspire more sophisticated PR methods.

\bibliography{aaai2027}


\end{document}